\documentclass[11pt,a4paper]{article}
\usepackage[hyperref]{acl2018}
\usepackage{times}
\usepackage{latexsym}
 \usepackage{multirow}
\usepackage{amsmath,bm}
\usepackage{hhline}
 \usepackage{soul}
\usepackage{slashbox}

\usepackage{times} 
\usepackage{helvet} 
\usepackage{courier} 
\usepackage{url} 
\usepackage{graphicx} 
\usepackage{times}
\usepackage{latexsym}
\usepackage{subcaption}
 
 \usepackage{amsmath}
\usepackage{amssymb}
\usepackage{tikz}
\usepackage{caption}
\usepackage{tabularx}
\usepackage{bm}
\usepackage[utf8]{inputenc}
 
\usepackage{url}

\aclfinalcopy 
\def\aclpaperid{1460} 

 \title{HotFlip: White-Box Adversarial Examples for Text Classification}

\author
       {Javid Ebrahimi$^*$, Anyi Rao$^\dagger$, Daniel Lowd$^*$, 
        Dejing Dou$^*$
       \\
       $^*$Computer and Information Science Department, University of Oregon, USA\\
       \{\texttt{javid, lowd, dou}\}@cs.uoregon.edu\\ 
       $^\dagger$School of Electronic Science and Engineering, Nanjing University, China\\
         \{\texttt{anyirao}\}@smail.nju.edu.cn\\ 
       }

\date{}

\begin{document}
\maketitle
\begin{abstract}


 We propose an efficient method to generate white-box  adversarial examples to trick a character-level neural classifier. We find that only a few manipulations are needed to greatly decrease the accuracy. 
 Our method relies on an atomic flip operation, which swaps one token for another, based on the gradients of the one-hot input vectors.
Due to efficiency of our method, we can perform adversarial training which makes the model more robust to attacks at test time. 
With the use of a few semantics-preserving constraints, we demonstrate that HotFlip can be adapted to attack a word-level classifier as well.


\end{abstract}

\section{Introduction}

Adversarial examples are inputs to a predictive machine learning model that are maliciously designed to cause poor performance \cite{goodfellow2014explaining}. Adversarial examples expose regions of the input space where the model performs poorly, which can aid in understanding and improving the model. By using these examples as training data, adversarial training learns models that are more robust, and may even perform better on non-adversarial examples. 
  Interest in understanding vulnerabilities of NLP systems is growing \cite{jia2017adversarial,zhao2017generating,belinkov2017synthetic,iyyer2018adversarial}. 
Previous work has focused on heuristics for creating adversarial examples in the \textit{black-box} setting, without any explicit knowledge of the model parameters.  In the \textit{white-box} setting, we use complete knowledge of the model to develop worst-case attacks, which can reveal much larger vulnerabilities.
 We propose a white-box adversary against differentiable text classifiers. We find that only a few manipulations are needed to greatly increase the misclassification error. Furthermore, fast generation of adversarial examples allows feasible adversarial training, which helps the model defend against adversarial examples and improve accuracy on clean examples. 
At the core of our method lies an atomic {\em flip} operation, which changes one token to another by using the directional derivatives of the model with respect to the one-hot vector input.

\begin{table}
\centering
\scalebox{0.8}{
\begin{tabularx}{0.45 \textwidth}{X}
\hline
\small{{South Africa's historic Soweto township marks its 100th birthday on Tuesday in a mood of optimism.} 57\% \textbf{World}}\\
\small{{South Africa's historic Soweto township marks its 100th birthday on Tuesday in a moo\textbf{P} of optimism.} 95\% \textbf{Sci/Tech}}\\
\hline
\small{{ Chancellor Gordon Brown has sought to quell speculation over who should run the Labour Party and turned the attack on the opposition Conservatives.} 75\% \textbf{World}}\\
\small{{Chancellor Gordon Brown has sought to quell speculation over who should run the Labour Party and turned the attack on the o\textbf{B}position Conservatives.} 94\% \textbf{Business}}\\

\hline
\end{tabularx}
}
\captionsetup{font=footnotesize}
\caption{Adversarial examples with a single character change, which will be misclassified by a neural classifier.}
\label{exmp}
\end{table}

Our contributions are as follows:
\begin{enumerate}
\item We propose an efficient gradient-based optimization method to manipulate discrete text structure at its one-hot representation.
\item We investigate the robustness of a classifier trained with adversarial examples, by studying its resilience to attacks and its accuracy on clean test data.
\end{enumerate}

\section{Related Work}
Adversarial examples are powerful tools to investigate the vulnerabilities of a deep learning model \cite{szegedy2013intriguing}.
While this line of research has recently received a lot of attention in the deep learning community, it has a long history in machine learning, going back to adversarial attacks on linear spam classifiers \cite{dalvi2004adversarial,lowd2005adversarial}.
 Hosseini et al. \shortcite{hosseini2017deceiving} show that simple modifications, such as adding spaces or dots between characters,  can drastically change the toxicity score from Google's \texttt{perspective} API \footnote{https://www.perspectiveapi.com}. 
Belinkov and Bisk \shortcite{belinkov2017synthetic} show that character-level machine translation systems are overly sensitive to random character manipulations, such as keyboard typos. They manipulate every word in a sentence with synthetic or natural noise. However, throughout our experiments, we care about the degree of distortion in a sentence, and look for stronger adversaries which can increase the loss within a limited budget. Instead of randomly perturbing text, we propose an efficient method, which can generate adversarial text using the gradients of the model with respect to the input. 






Adversarial training interleaves training with generation of adversarial examples \cite{goodfellow2014explaining}. 
Concretely, after every iteration of training, adversarial examples are created and added to the mini-batches. 
A projected gradient-based approach to create adversarial examples by Madry et al. \shortcite{madry2017towards} has proved to be one of the most effective defense mechanisms against adversarial attacks for image classification.
Miyato et al. \shortcite{miyato2016adversarial} {create adversarial examples by adding noise to word embeddings, without creating real-world textual adversarial examples.  Our work is the first to propose an efficient method to generate real-world adversarial examples which can also be used for effective adversarial training.}

\section{HotFlip}
%
 HotFlip is a method for generating adversarial examples with character substitutions (``flips''). HotFlip also supports insertion and deletion operations by representing them as sequences of character substitutions.  It uses the gradient with respect to a one-hot input representation to efficiently estimate which individual change has the highest estimated loss, and it uses a beam search to find a set of manipulations that work well together to confuse a classifier.
\subsection{Definitions}
We use $J(\mathbf{x}, \mathbf{y})$ to refer to the loss of the model on input $\mathbf{x}$ with true output $\mathbf{y}$. For example, for classification, the loss would be the log-loss over the output of the softmax unit.
Let $V$ be the alphabet, $\mathbf{x}$ be a text of length $L$ characters, and $x_{ij} \in \{0,1\}^{|V|}$ denote a one-hot vector representing the $j$-th character of the $i$-th word. The character sequence can be represented by
\centerline{$\mathbf{x}$ = [($x_{11}$,.. $x_{1n}$);..($x_{m1}$,.. $x_{mn}$)]}
wherein a semicolon denotes explicit segmentation between words. The number of words is denoted by $m$, and $n$ is the number of maximum characters allowed for a word.

\subsection{Derivatives of Operations}

We represent text operations as vectors in the input space and estimate the change in loss by directional derivatives with respect to these operations. Based on these derivatives, the adversary can choose the best loss-increasing direction. Our algorithm requires just one function evaluation (forward pass) and one gradient computation (backward pass) to estimate the best possible flip. 

A  \textbf{flip} of the $j$-th character of the $i$-th word ($a \rightarrow b$) can be represented by this vector:
\\\\
\centerline{
$\vec{v}^{\,}_{ijb}$ = ($\vec{0}^{\,}$,..;$($$\vec{0}^{\,}$,..$($0,..-1,0,..,1,0$)_j$,..$\vec{0}^{\,}$$)_i$; $\vec{0}^{\,}$,..)
}\\\\
where -1 and 1 are in the corresponding positions for the $a$-th and $b$-th characters of the alphabet, respectively, and $x_{ij}^{(a)}=1$. A first-order approximation of change in loss can be obtained from a directional derivative along this vector: 
\\\\
\centerline{
$\nabla_{\vec{v}^{\,}_{ijb}}J(\mathbf{x}, \mathbf{y}) = \nabla_{x}J(\mathbf{x}, \mathbf{y})^{T} \cdot \> \vec{v}^{\,}_{ijb}$
}
\\\\
We choose the vector with biggest increase in loss:
\begin{equation}
\begin{split}
\underset{}{\text{max}} \nabla_{x}J(\mathbf{x}, \mathbf{y})^{T} \cdot \> \vec{v}^{\,}_{ijb}= \underset{{ijb}}{\text{max}} {{\frac{\partial{J}}{\partial{x_{ij}}}}^{(b)} - { \frac{\partial{J}}{\partial{x_{ij}}}}^{(a)}}
\end{split}
\label{eq1}
\end{equation}
Using the derivatives as a surrogate loss, we simply need to find the best change by calling the function mentioned in eq. \ref{eq1}, to \textit{estimate} the best character change ($a \rightarrow b$).  
This is in contrast to a naive loss-based approach, which has to query the classifier for every possible change to compute the \textit{exact} loss induced by those changes. In other words, apart from the overhead of calling the function in eq. \ref{eq1}, one backward pass saves the adversary a large number of forward passes.

Character \textbf{insertion}\footnote{For ease in exposition, we assume that the word size is at most $n$-1, leaving at least one position of padding at the end.} at the $j$-th position of the $i$-th word can also be treated as a character flip, followed by more flips as characters are shifted to the right until the end of the word.
\begin{multline*}
\underset{}{\text{max}} \nabla_{x}J(\mathbf{x}, \mathbf{y})^{T} \cdot \> \vec{v}^{\,}_{ijb}= \underset{{ijb}}{\text{max}} {{\frac{\partial{J}}{\partial{x_{ij}}}}^{(b)} - { \frac{\partial{J}}{\partial{x_{ij}}}}^{(a)}} \\
+ \sum_{j^{'}=j+1}^n \bigg( {{\frac{\partial{J}}{\partial{x_{ij^{'}}}}}^{(b^{'})} - { \frac{\partial{J}}{\partial{x_{ij^{'}}}}}^{(a^{'})}} \bigg)
\label{eq-der}
\end{multline*}
where $x_{ij^{'}}^{(a^{'})}=1$ and $x_{i{j^{'}{-1}}}^{(b^{'})}=1$.
Similarly, character \textbf{deletion} can be written as a number of character flips as characters are shifted to the left. 
Since the magnitudes of direction vectors (operations) are different, we normalize by the $L_2$ norm of the vector i.e., $\frac{\vec{v}}{\sqrt{2N}}$, where N is the number of total flips.


\subsection{Multiple Changes}
We explained how to estimate the best single change in text to get the maximum increase in loss. A greedy or beam search of $r$ steps will give us an adversarial example with a maximum of $r$ flips, or more concretely an adversarial example within an $L_0$ distance of $r$ from the original example. 
Our beam search requires only $\mathcal{O}(br)$ forward passes and an equal number of backward passes, with $r$ being the budget and $b$, the beam width.
We elaborate on this with an example: Consider the loss function $J(.)$, input $x_0$, and an individual change $c_j$. We estimate the score for the change as $\frac{\partial{J(x_0)}}{\partial{c_j}}$.
For a sequence of 3 changes [$c_1$,$c_2$,$c_3$], we evaluate the ``score'' as follows.
\begin{equation*}
\text{score([}c_1,c_2,c_3\text{])} = \frac{\partial{J(x_0)}}{\partial{c_1}} + \frac{\partial{J(x_1)}}{\partial{c_2}} + \frac{\partial{J(x_2)}}{\partial{c_3}}
\end{equation*}
where $x_1$ and $x_2$ are the modified input after applying [$c_1$] and [$c_1$, $c_2$] respectively. 
We need $b$ forward and backward passes to compute derivatives at each step of the path, leading to $\mathcal{O}(br)$ queries. In contrast, a naive loss-based approach requires computing the exact loss for every possible change at every stage of the beam search, leading to $\mathcal{O}(brL|V|)$ queries.
\section{Experiments}

\begin{figure}
\centering
\scalebox{0.8}{
\includegraphics[width=\linewidth]{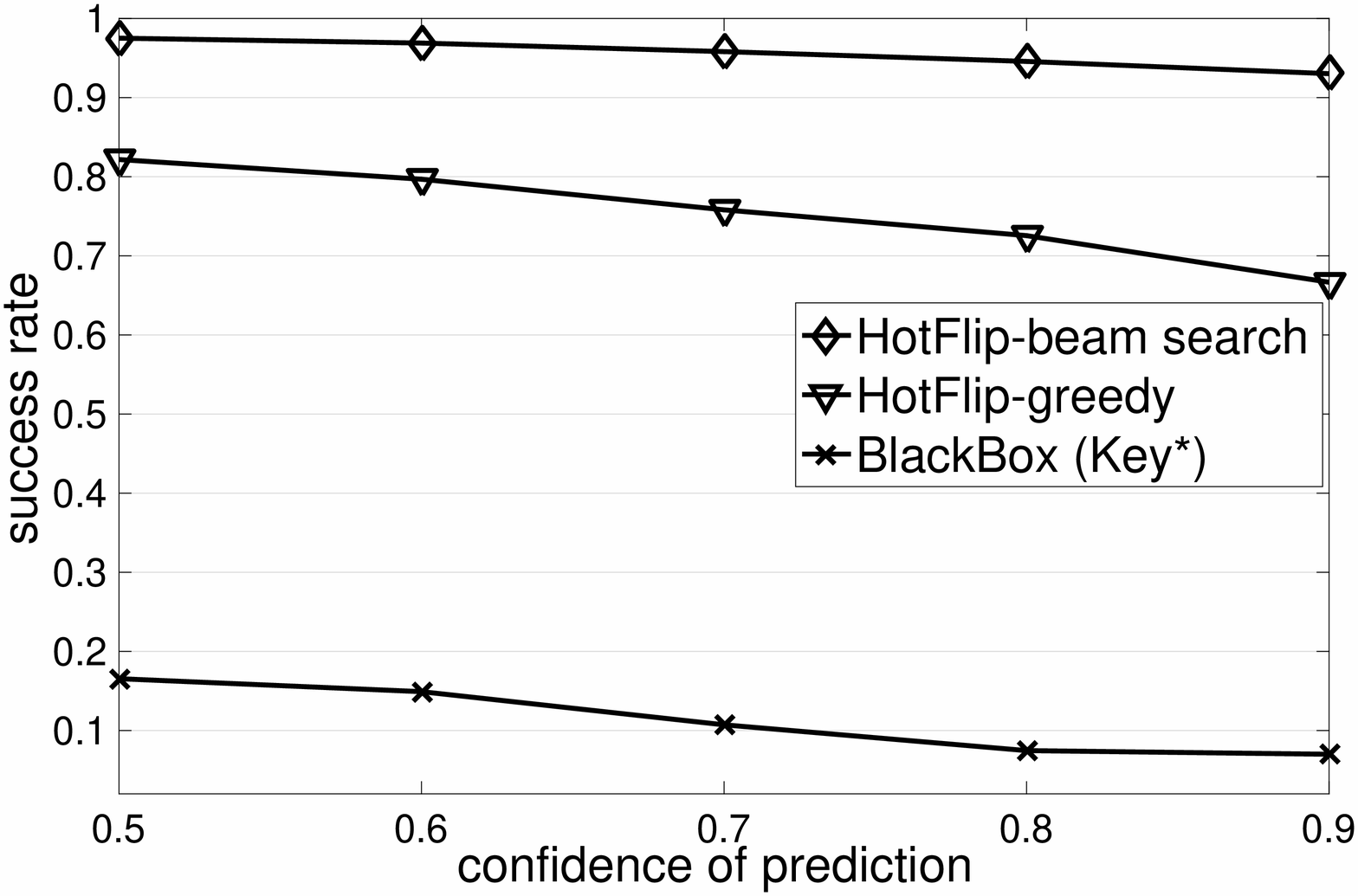}
}
\captionsetup{font=footnotesize}

\caption{Adversary's success as a function of confidence.}\label{fig:successcl}
\end{figure}

 In principle, HotFlip could be applied to any differentiable character-based classifier. Here, we focus on the CharCNN-LSTM architecture \cite{kim2015character}, which can be adapted for classification via a single dense layer after the last recurrent hidden unit.
We use the {AG's news} dataset\footnote{https://www.di.unipi.it/\textasciitilde gulli/}, which consists of 120,000 training and 7,600 test instances from four equal-sized classes: World, Sports, Business, and Science/Technology.
The architecture consists of a 2-layer stacked LSTM with 500 hidden units, a character embedding size of 25, and 1000 kernels of width 6 for temporal convolutions. This classifier was able to outperform \cite{conneau2017very}, which has achieved the state-of-the-art result on some benchmarks, on AG's news. 
The model is trained with SGD and gradient clipping, and the batch size was set to 64.
We used 10\% of the training data as the development set, and trained for a maximum of 25 epochs. 
We only allow character changes if the new word does not exist in the vocabulary, to avoid changes that are more likely to change the meaning of text. The adversary uses a beam size of 10, and has a budget of maximum of 10\% of characters in the document. 
In Figure \ref{fig:successcl}, we plot the success rate of the adversary against an acceptable confidence score for the misclassification. That is, we consider the adversary successful only if the classifier misclassifies the instance with a given confidence score. For this experiment, we create adversarial examples for 10\% of the test set.

We compare with a (greedy) black-box adversary, which does not have access to model parameters, and simply queries the classifier with random character changes. Belinkov and Bisk \shortcite{belinkov2017synthetic} define an attack, \texttt{Key}, in which a character is replaced with an adjacent character in the keyboard. We allow a stronger black-box attacker to change a character to any character in the alphabet, and we call it \texttt{Key}$^*$. 
As expected a white-box adversary is much more damaging, and has a higher success rate. As can be seen, the beam-search strategy is very effective in fooling the classifier even with an 0.9 confidence constraint, tricking the classifier for more than 90\% of the instances. 
A greedy search is less effective especially in producing high-confidence scores. 
 We use a maximum of 10\% of characters in the document as the budget for the adversary, but our adversary changes an average of 4.18\% of the characters to trick the classifier at confidence 0.5. The adversary picks the flip operation around 80\% of the times, and favors delete over insert by two to one.   

%
%
%

%
%

\subsection{Robustness}
%
%
%
For our adversarial training, we use only use the {flip} operation, and evaluate models' robustness to this operation only. This is because insert and delete manipulations are $n$ times slower to generate, where $n$ is the number of maximum characters allowed for a word. For these experiments, we have no constraint on confidence score. We flip $r$ characters for each training sample, which was set to
20\% of the characters in text after tuning, based on the accuracy on the development set. In addition, for faster generation of adversarial examples, we directly apply the top $r$ flips after the first backward pass, simultaneously\footnote{The adversary at test time would still use beam search.}. 

We use the full test set for this experiment, and we compare HotFlip adversarial training with the white-box (supervised) adversarial training \cite{miyato2016adversarial} that perturbs word embeddings, which we adapt to work with character embeddings. Specifically, the adversarial noise per character is constrained by the Frobenius norm of the embedding matrix composed of the sequence of characters in the word. 
  We also create another baseline where instead of white-box adversarial examples, we add black-box adversarial examples ($\texttt{Key}^{*}$) to the mini-batches. As shown in Table \ref{tablereg}, our approach decreases misclassification error and
dramatically decreases the adversary's success rate. In particular, adversarial training on real adversarial examples generated by HotFlip, is more effective than training on \textit{pseudo}-adversarial examples created by adding noise to the embeddings. 

The current error of our adversarially trained model is still beyond an acceptable rate; this is mainly because the adversary that we use at test time, which uses beam search, is strictly stronger than our model's internal adversary. This has been observed in computer vision where strongest adversaries are not efficient enough for adversarial training, but can break models trained with weaker adversaries \cite{carlini2017towards}.


\begin{table}
\centering
\scalebox {0.8} {
\begin{tabular}{c | c | c }
Method & Misc. error  & Success rate \\
\hline
Baseline & 8.27\% & 98.16\%  \\ 
\hline
Adv-tr \cite{miyato2016adversarial} & 8.03\% & 87.43\% \\ 
\hline
Adv-tr (black-box)  & {8.60}\%  &  {95.63}\% \\
\hline
Adv-tr (white-box) & \textbf{7.65}\%  &  \textbf{69.32}\% \\ 
\hline
\end{tabular}
}
\captionsetup{font=footnotesize}
\caption{Comparison based on misclassification error on clean data and adversary's success rate.
}
\label{tablereg}
\end{table}

\subsection{Human Perception}
Our human evaluation experiment shows that our character-based adversarial examples rarely alter the meaning of a sentence. We conduct an experiment of annotating 600 randomly-picked instances annotated by at least three crowd workers in Amazon Mechanical Turk. This set contains 150 examples of each class of AG's-news dataset, all of which are correctly classified by the classifier. We manipulate half of this set by our algorithm, which can successfully trick the classifier to misclassify these 300 adversarial examples. 
The median accuracy of our participants decreased by 1.78\% from 87.49\% on clean examples to 85.71\% on adversarial examples.
 Similar small drops in human performance have been reported for image classification \cite{papernot2016limitations} and text comprehension \cite{jia2017adversarial}. 

\section{HotFlip at Word-Level}
\begin{table*}
\centering
\scalebox {0.7} {

\begin{tabular}{p{18cm}  }
\hline
one hour photo is an intriguing (\textbf{interesting}) snapshot of one man and his delusions it's just too bad it doesn't have more flashes of insight.\\
\hline
`enigma' is a good (\textbf{terrific}) name for a movie this deliberately obtuse and unapproachable.\\
\hline
an intermittently pleasing (\textbf{satisfying})  but mostly routine effort.\\
\hline
an atonal estrogen opera that demonizes feminism while gifting the most sympathetic male of the piece with a nice (\textbf{wonderful}) vomit bath at his wedding.\\
\hline
culkin exudes (\textbf{infuses}) none of the charm or charisma that might keep a more general audience even vaguely interested in his bratty character.
\end{tabular}
}
\captionsetup{font=footnotesize}
\caption{Adversarial examples for sentiment classification. The bold words replace the words before them.}
\label{tableword}
\end{table*}

\begin{table*}
\center
\scalebox {0.65} {
\begin{tabular}{l | l | l | l | l | l | l | l}
past $\rightarrow$ pas!t & Alps $\rightarrow$ llps & talk $\rightarrow$ taln & local $\rightarrow$ loral & you $\rightarrow$ yoTu & ships $\rightarrow$ hips & actor $\rightarrow$ actr & lowered $\rightarrow$ owered\\
\hline
pasturing & lips & tall & moral & Tutu & dips & act & powered\\
pasture & laps & tale & Moral & Hutu & hops & acting & empowered\\
pastor & legs & tales & coral & Turku & lips & actress & owed\\
Task & slips & talent & morals & Futurum & hits & acts & overpowered \\

\end{tabular}
}
\captionsetup{font=footnotesize}
\caption{Nearest neighbor words (based on cosine similarity) of word representations from CharCNN-LSTM, picked at the output of the highway layers. A single adversarial change in the word often results in a big change in the embedding, which would make the word more similar to other words, rather than to the original word.
}
\label{table_emb}
\end{table*}

HotFlip can naturally be adapted to generate adversarial examples for word-level models, by computing derivatives with respect to one-hot word vectors.
After a few character changes, the meaning of the text is very likely to be preserved or inferred by the reader \cite{rawlinson1976significance}, which was also confirmed by our human subjects study. 
By contrast, word-level adversarial manipulations are much more likely to change the meaning of text, which makes the use of semantics-preserving constraints necessary. 
For example, changing the word \textit{good} to \textit{bad} changes the sentiment of the sentence ``\textit{this was a good movie}''. In fact, we expect the model to predict a different label after such a change.

 To showcase the applicability of HotFlip to a word-level classifier, we use Kim's CNN \shortcite{kim2014convolutional} trained for binary sentiment classification on the SST dataset~\cite{socher2013recursive}. 
 In order to create {adversarial} examples, we add constraints so that the resulting sentence is likely to preserve the original meaning; we only flip a word $w_i$ to $w_j$ only if these constraints are satisfied: 
\begin{enumerate}
\item 
The cosine similarity between the embedding of words is bigger than a threshold (0.8).
\item 
The two words have the same part-of-speech. 
\item 
We disallow replacing of stop-words, as for many of the stop-words, it is difficult to find cases where replacing them will still render the sentence grammatically correct. We also disallow changing a word to another word with the same lexeme for the same purpose. 
\end{enumerate}

Table \ref{tableword} shows a few adversarial examples with only one word flip. In the second and the fourth examples, the adversary flips a positive word (i.e., \textit{good}, \textit{nice}) with highly positive words (i.e., \textit{terrific}, \textit{wonderful}) in an overall very negative review. 
These examples, albeit interesting and intuitive, are not abundant, and thus pose less threat to an NLP word-level model. Specifically, given the strict set of constraints, we were able to create only 41 examples (2\% of the correctly-classified instances of the SST test set) with one or two flips. 

For a qualitative analysis of relative brittleness of character-level models, we study the change in word embedding as an adversarial flip, insert, or delete operation occurs in Table \ref{table_emb}. We use the output of the highway layer as the word representation, and report the embedding for a few adversarial words, for which the original word is not among their top 5 nearest neighbors. 

In a character-level model, the lookup operation to pick a word from the vocabulary is replaced by a character-sequence feature extractor which gives an embedding for any input, including OOV words which would be mapped to an UNK token in a word-level model. This makes the embedding space induced in character-level representation more dense, which makes character-level models more likely to misbehave under small adversarial perturbations.  

\section{Conclusion and Future Work}
White-box attacks are among the most serious forms of attacks an adversary can inflict on a machine learning model. 
We create white-box adversarial examples by computing derivatives with respect to a few character-edit operations (i.e., flip, insert, delete), which can be used in a beam-search optimization.
While character-edit operations have little impact on human understanding, we found that character-level models are highly sensitive to adversarial perturbations. 
Employing these adversarial examples in adversarial training renders the models more robust to such attacks, as well as more robust to unseen clean data. 

 
Contrasting and evaluating robustness of different character-level models for different tasks is an important future direction for adversarial NLP. In addition, the discrete nature of text makes it a more challenging task to understand the landscape of adversarial examples. Research in this direction can shed light on vulnerabilities of NLP models. 
\section{Acknowledgement}
This work was funded by ARO grant W911NF-15-1-0265.

\bibliography{acl2018}

\begin{thebibliography}{18}
\expandafter\ifx\csname natexlab\endcsname\relax\def\natexlab#1{#1}\fi

\bibitem[{Belinkov and Bisk(2018)}]{belinkov2017synthetic}
Yonatan Belinkov and Yonatan Bisk. 2018.
\newblock Synthetic and natural noise both break neural machine translation.
\newblock In \emph{{Proceedings of ICLR}}.

\bibitem[{Carlini and Wagner(2017)}]{carlini2017towards}
Nicholas Carlini and David Wagner. 2017.
\newblock Towards evaluating the robustness of neural networks.
\newblock In \emph{Security and Privacy (SP), 2017 IEEE Symposium on}, pages
  39--57. IEEE.

\bibitem[{Conneau et~al.(2017)Conneau, Schwenk, Barrault, and
  Lecun}]{conneau2017very}
Alexis Conneau, Holger Schwenk, Lo{\"\i}c Barrault, and Yann Lecun. 2017.
\newblock Very deep convolutional networks for text classification.
\newblock In \emph{Proceedings of {EACL}}, volume~1, pages 1107--1116.

\bibitem[{Dalvi et~al.(2004)Dalvi, Domingos, Sanghai, Verma
  et~al.}]{dalvi2004adversarial}
Nilesh Dalvi, Pedro Domingos, Sumit Sanghai, Deepak Verma, et~al. 2004.
\newblock Adversarial classification.
\newblock In \emph{Proceedings of KDD}, pages 99--108.

\bibitem[{Goodfellow et~al.(2015)Goodfellow, Shlens, and
  Szegedy}]{goodfellow2014explaining}
Ian~J Goodfellow, Jonathon Shlens, and Christian Szegedy. 2015.
\newblock Explaining and harnessing adversarial examples.
\newblock In \emph{Proceedings of ICLR}.

\bibitem[{Hosseini et~al.(2017)Hosseini, Kannan, Zhang, and
  Poovendran}]{hosseini2017deceiving}
Hossein Hosseini, Sreeram Kannan, Baosen Zhang, and Radha Poovendran. 2017.
\newblock Deceiving google's perspective api built for detecting toxic
  comments.
\newblock \emph{arXiv preprint arXiv:1702.08138}.

\bibitem[{Iyyer et~al.(2018)Iyyer, Wieting, Gimpel, and
  Zettlemoyer}]{iyyer2018adversarial}
Mohit Iyyer, John Wieting, Kevin Gimpel, and Luke Zettlemoyer. 2018.
\newblock Adversarial example generation with syntactically controlled
  paraphrase networks.
\newblock \emph{arXiv preprint arXiv:1804.06059}.

\bibitem[{Jia and Liang(2017)}]{jia2017adversarial}
Robin Jia and Percy Liang. 2017.
\newblock Adversarial examples for evaluating reading comprehension systems.
\newblock In \emph{Proceedings of EMNLP}.

\bibitem[{Kim(2014)}]{kim2014convolutional}
Yoon Kim. 2014.
\newblock Convolutional neural networks for sentence classification.
\newblock In \emph{Proceedings of EMNLP}.

\bibitem[{Kim et~al.(2016)Kim, Jernite, Sontag, and Rush}]{kim2015character}
Yoon Kim, Yacine Jernite, David Sontag, and Alexander~M Rush. 2016.
\newblock Character-aware neural language models.
\newblock \emph{Proceedings of AAAI}.

\bibitem[{Lowd and Meek(2005)}]{lowd2005adversarial}
Daniel Lowd and Christopher Meek. 2005.
\newblock Adversarial learning.
\newblock In \emph{Proceedings of KDD}, pages 641--647.

\bibitem[{Madry et~al.(2018)Madry, Makelov, Schmidt, Tsipras, and
  Vladu}]{madry2017towards}
Aleksander Madry, Aleksandar Makelov, Ludwig Schmidt, Dimitris Tsipras, and
  Adrian Vladu. 2018.
\newblock Towards deep learning models resistant to adversarial attacks.
\newblock In \emph{Proceedings of ICLR}.

\bibitem[{Miyato et~al.(2017)Miyato, Dai, and
  Goodfellow}]{miyato2016adversarial}
Takeru Miyato, Andrew~M Dai, and Ian Goodfellow. 2017.
\newblock Adversarial training methods for semi-supervised text classification.
\newblock In \emph{Proceedings of ICLR}.

\bibitem[{Papernot et~al.(2016)Papernot, McDaniel, Jha, Fredrikson, Celik, and
  Swami}]{papernot2016limitations}
Nicolas Papernot, Patrick McDaniel, Somesh Jha, Matt Fredrikson, Z~Berkay
  Celik, and Ananthram Swami. 2016.
\newblock The limitations of deep learning in adversarial settings.
\newblock In \emph{Security and Privacy (EuroS\&P), 2016 IEEE European
  Symposium on}, pages 372--387. IEEE.

\bibitem[{Rawlinson(1976)}]{rawlinson1976significance}
Graham~Ernest Rawlinson. 1976.
\newblock \emph{The Significance of Letter Position in Word Recognition.}
\newblock Ph.D. thesis, University of Nottingham.

\bibitem[{Socher et~al.(2013)Socher, Perelygin, Wu, Chuang, Manning, Ng, and
  Potts}]{socher2013recursive}
Richard Socher, Alex Perelygin, Jean Wu, Jason Chuang, Christopher~D Manning,
  Andrew Ng, and Christopher Potts. 2013.
\newblock Recursive deep models for semantic compositionality over a sentiment
  treebank.
\newblock In \emph{Proceedings of EMNLP}, pages 1631--1642.

\bibitem[{Szegedy et~al.(2014)Szegedy, Zaremba, Sutskever, Bruna, Erhan,
  Goodfellow, and Fergus}]{szegedy2013intriguing}
Christian Szegedy, Wojciech Zaremba, Ilya Sutskever, Joan Bruna, Dumitru Erhan,
  Ian Goodfellow, and Rob Fergus. 2014.
\newblock Intriguing properties of neural networks.
\newblock In \emph{Proceedings of ICLR}.

\bibitem[{Zhao et~al.(2018)Zhao, Dua, and Singh}]{zhao2017generating}
Zhengli Zhao, Dheeru Dua, and Sameer Singh. 2018.
\newblock Generating natural adversarial examples.
\newblock In \emph{{Proceedings of ICLR}}.

\end{thebibliography}
\bibliographystyle{acl_natbib}

\end{document}